\documentclass{article}

\usepackage{microtype}
\usepackage{graphicx}
\usepackage{subcaption}
\usepackage{booktabs} %

\usepackage{hyperref}

\usepackage[accepted]{icml2026_weightsymmetry}

\usepackage{amsmath}
\usepackage{amssymb}
\usepackage{mathtools}
\usepackage{amsthm}

\usepackage[capitalize,noabbrev]{cleveref}

\theoremstyle{plain}

\theoremstyle{definition}

\theoremstyle{remark}

\usepackage[textsize=tiny]{todonotes}

\icmltitlerunning{How the Optimizer Shapes Learned Solutions in Equivariant Neural Networks}

\begin{document}

\twocolumn[
  \icmltitle{How the Optimizer Shapes Learned Solutions in Equivariant Neural Networks}

  \icmlsetsymbol{equal}{*}

  \begin{icmlauthorlist}
    \icmlauthor{Teodor-Mihai Stupariu}{bdf,cni}
    \icmlauthor{Andrei Manolache}{bdf,ustr,mpi}
    \end{icmlauthorlist}
    
    \icmlaffiliation{bdf}{Bitdefender, Romania}
    \icmlaffiliation{ustr}{University of Stuttgart, Germany}
    \icmlaffiliation{mpi}{International Max Planck Research School for Intelligent Systems, Germany}
    \icmlaffiliation{cni}{Tudor Vianu High School of Computer Science, Romania}

  \icmlcorrespondingauthor{Andrei Manolache}{andrei.manolache@ki.uni-stuttgart.de}

  \icmlkeywords{Machine Learning, ICML}

  \vskip 0.3in
]

\newcommand{\am}[1]{{{\textcolor{orange}{\textbf{[AM:} {#1}\textbf{]}}}}}  %

\printAffiliationsAndNotice{}  %

\begin{abstract}

Equivariant neural networks encode geometric symmetries by construction, yet they are often difficult to optimize and can underperform less constrained architectures. A growing body of work addresses this through architectural modifications such as constraint relaxation or approximate equivariance, while the role of the optimizer remains comparatively underexplored. We study this direction by comparing Muon and Adam across several equivariant and geometric architectures under pointcloud and molecular learning settings. On ModelNet40, where the comparison is clearest, Muon consistently improves over Adam across all architectures considered. We then analyze the trained ModelNet40 checkpoints through Hessian estimates, loss surface visualizations, and spectral properties of learned weights and intermediate representations. The checkpoints reached by Muon have larger Hessian curvature summaries but more regular loss surfaces, and their learned weights and representations have higher stable and effective ranks. These observations suggest that the interaction between optimizer design and geometric inductive bias deserves further attention from the community. 

\end{abstract}

\section{Introduction and Related Work}

Equivariant neural networks encode geometric symmetries directly into their architecture, providing a natural inductive bias for problems whose data carries known geometric symmetries
~\citep{thomas18, pmlr-v139-satorras21a, bronstein21}. Hard equivariance constraints make optimization difficult, with the loss landscape of equivariant models shown to contain critical points and even spurious minima~\citep{xie2025a}, and can underperform less constrained alternatives at scale~\citep{xie2025the, brehmer2025doesequivariancematterscale}. A natural response, and the dominant one in recent literature, has been to soften the architectural constraints. Approximate~\citep{wang2022approximately, petrache2023approximationgeneralization} and relaxed equivariance~\citep{pertigkiozoglou2024improving, manolache2025learning, elhag2025relaxed} methods allow controlled departures from exact symmetry, improving optimization while retaining some symmetry bias. These methods place the architectural constraints at the center of the optimization problem, while the optimizer has received almost no attention in this literature.

Recent work argues that the optimizer is a meaningful source of inductive bias, and that different optimizers do not merely converge at different rates but arrive at qualitatively different solutions when training neural networks~\citep{pascanu2025optimizersqualitativelyaltersolutions}. If the difficulty of training equivariant models is partly an optimization problem~\citep{pertigkiozoglou2024improving, manolache2025learning}, then changing the optimizer is a natural place to intervene, and one largely independent of the architectural relaxations above.
We test this idea empirically by comparing Adam~\cite{adam} with Muon~\cite{jordan2024muon}, a recently proposed optimizer that orthogonalizes its momentum buffer through Newton-Schulz iterations. Our primary setting is ModelNet40~\cite{Wu20143DSA} and its corrupted variant ModelNet40-C~\citep{sun2022benchmarking}, where we compare architectures with different geometric inductive biases: EGNN~\cite{pmlr-v139-satorras21a}, with explicit $E(n)$ equivariance; DGCNN~\citep{dgcnn}, a permutation-equivariant graph convolutional network that builds local geometric structure through dynamic k-NN connectivity; and PointNet~\citep{qi2016pointnet}, which is permutation invariant via symmetric pooling. We complement this comparison with molecular-learning experiments using the $E(3)$-equivariant GotenNet~\citep{aykent2025gotennet} and the permutation-equivariant GINE~\citep{Hu2020Strategies} on molecular datasets~\citep{Ramakrishnan2014Aug, dwivedi2022long, Gomez-Bombarelli2018Feb}.

On ModelNet40, where the comparison is clearest, Muon and Adam differ beyond test accuracy. The trained solutions show distinct local loss geometry and distinct spectral structure in their learned weights and intermediate representations. Muon also reaches higher-curvature checkpoints but with surrounding loss structure that appears more regular. These observations motivate treating optimizer choice as a meaningful part of training equivariant models.

We summarize our contributions below:

\begin{enumerate}
    \item We compare Adam and Muon across equivariant and geometric architectures under clean, corrupted, and molecular-learning setups.
    \item We study the local loss geometry using Hessian estimates and loss-surface visualizations, finding higher curvature for Muon but smoother local loss structure.
    \item We examine the spectral structure of learned weights and intermediate representations, finding higher stable and effective ranks under Muon.
\end{enumerate}

\section{Experiments and Analysis}

We organize the experiments around three questions. First, \textit{how does replacing Adam with Muon affect performance across architectures and evaluation settings?} Second, \textit{how do the resulting solutions differ in local loss geometry?} Third, \textit{does optimizer choice affect the spectral structure of the learned weights and representations?} We begin with the quantitative comparison on ModelNet40, its corrupted variant, and molecular-learning benchmarks.

\subsection{Quantitative evaluation}

\begin{table}[t]
\caption{Best-checkpoint accuracy on ModelNet40 (``Clean'') and ModelNet40-C (``Corrupted''). Results are mean accuracy $\pm$ std over four seeds; $\Delta$ is Muon minus Adam. Higher is better. Muon improves over Adam for every model and evaluation setting.}
\label{tab:modelnet40}
\centering
{\setlength{\tabcolsep}{3.2pt}
\begin{tabular}{@{}llccc@{}}
\toprule
Eval. & Model & Adam & Muon & $\Delta$ \\
\midrule
Clean & EGNN     & 76.91{\tiny$\pm$0.94} & \textbf{82.08{\tiny$\pm$0.36}} & +5.17 \\
Clean & PointNet & 84.53{\tiny$\pm$0.70} & \textbf{87.21{\tiny$\pm$0.39}} & +2.67 \\
Clean & DGCNN    & 87.10{\tiny$\pm$0.69} & \textbf{89.06{\tiny$\pm$0.17}} & +1.96 \\
\midrule
Corrupted & EGNN     & 65.76{\tiny$\pm$0.95} & \textbf{70.12{\tiny$\pm$0.10}} & +4.36 \\
Corrupted & PointNet & 72.85{\tiny$\pm$1.05} & \textbf{75.87{\tiny$\pm$0.28}} & +3.02 \\
Corrupted & DGCNN    & 75.26{\tiny$\pm$1.63} & \textbf{77.84{\tiny$\pm$0.27}} & +2.58 \\
\bottomrule
\end{tabular}}
\end{table}

\begin{table}[t]
\caption{GotenNet performance on QM9. Results are mean MAE $\pm$ std over four seeds. Lower is better. Muon improves over Adam on most targets.}
\centering
\begin{tabular}{lcc}
\toprule
Property & Adam & Muon \\
\midrule
Cv    & \textbf{0.0230{\tiny$\pm$0.0008}} & 0.0246{\tiny$\pm$0.0015} \\
G     & 0.0103{\tiny$\pm$0.0032} & \textbf{0.0087{\tiny$\pm$0.0017}} \\
H     & 0.0122{\tiny$\pm$0.0020} & \textbf{0.0085{\tiny$\pm$0.0020}} \\
U     & \textbf{0.0101{\tiny$\pm$0.0042}} & 0.0138{\tiny$\pm$0.0079} \\
U0    & 0.0103{\tiny$\pm$0.0054} & \textbf{0.0090{\tiny$\pm$0.0022}} \\
zpve  & \textbf{0.0021{\tiny$\pm$0.0003}} & 0.0022{\tiny$\pm$0.0005} \\
mu    & 0.0124{\tiny$\pm$0.0014} & \textbf{0.0121{\tiny$\pm$0.0011}} \\
alpha & 0.0445{\tiny$\pm$0.0058} & \textbf{0.0408{\tiny$\pm$0.0043}} \\
homo  & 0.0225{\tiny$\pm$0.0016} & \textbf{0.0214{\tiny$\pm$0.0007}} \\
lumo  & 0.0177{\tiny$\pm$0.0019} & \textbf{0.0173{\tiny$\pm$0.0012}} \\
gap   & 0.0377{\tiny$\pm$0.0017} & \textbf{0.0363{\tiny$\pm$0.0025}} \\
r2    & 0.4320{\tiny$\pm$0.1222} & \textbf{0.2310{\tiny$\pm$0.1076}} \\
\bottomrule
\end{tabular}
\label{tab:qm9}
\end{table}
For each dataset and optimizer, we perform a grid search over learning rate and weight decay. We then train four seeds with the selected configuration and report best-checkpoint performance.
We first evaluate Adam and Muon on ModelNet40 using EGNN, PointNet, and DGCNN. These architectures impose different geometric inductive biases: EGNN enforces $E(3)$-equivariance, DGCNN builds local geometric structure through dynamic neighbourhoods, and PointNet relies primarily on permutation-invariant aggregation.

Table~\ref{tab:modelnet40} shows the main classification results. Muon improves over Adam for all three architectures on ModelNet40, and the same ordering holds under ModelNet40-C corruptions, showing clearly that changing only the optimizer can improve performance across architectures with different geometric biases. 

We next evaluate GotenNet on QM9~\citep{Ramakrishnan2014Aug}, where the task again involves $E(3)$-equivariant modelling. Table~\ref{tab:qm9} shows that Muon improves most QM9 targets. 
The additional (permutation-equivariant) message-passing experiments in Appendix~\ref{app:more_quantitative} are less conclusive. With GINE on Peptides-func~\citep{dwivedi2022long} (\cref{tab:pep-fun}) and ZINC~\citep{Gomez-Bombarelli2018Feb} (\cref{tab:zinc}), Muon reaches stronger best checkpoints on some seeds, but seed-averaged performance is comparable to Adam on Peptides-func and worse on ZINC.
These results suggest that optimizer choice is most consequential for some model-dataset pairs, with the strongest evidence in our experiments coming from 3D geometric settings.

\subsection{Local loss geometry}
\label{sec:loss-geom}

\begin{figure*}[t]
    \centering
    \vspace{-0.5em}
    \begin{minipage}{0.49\textwidth}
        \centering
        \includegraphics[width=\linewidth]{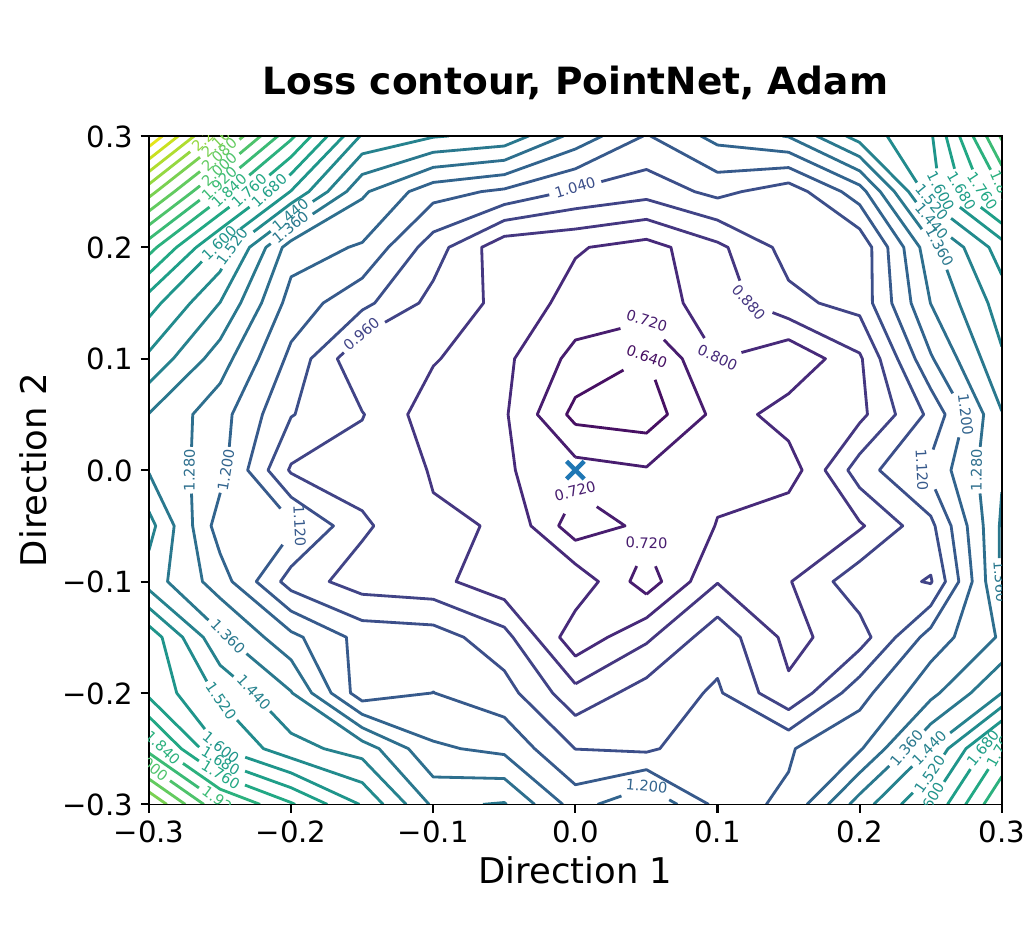}
    \end{minipage}
    \hfill
    \begin{minipage}{0.49\textwidth}
        \centering
        \includegraphics[width=\linewidth]{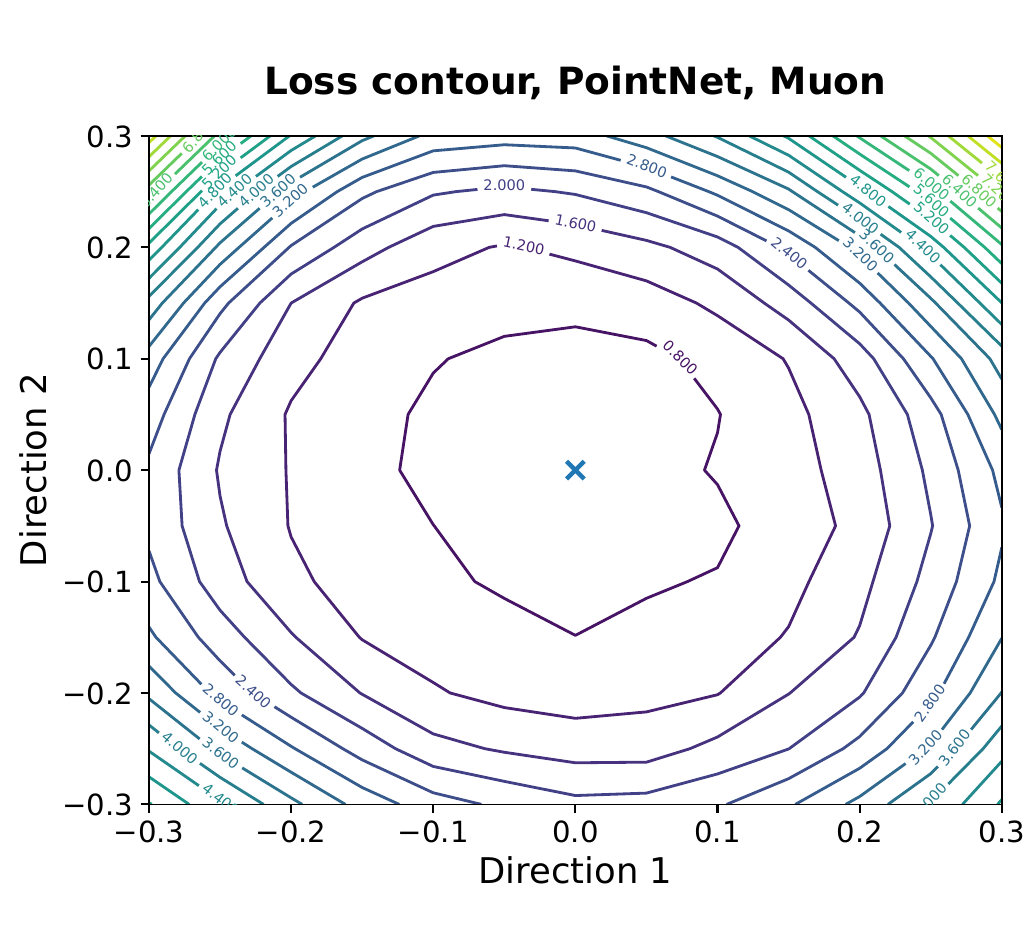}
    \end{minipage}
    \caption{Local loss contours around representative PointNet checkpoints trained with Adam and Muon on ModelNet40. The plots show normalized two-dimensional loss slices around each trained solution; the Muon slice appears smoother, while the Adam slice shows more irregular contour structure.}
    \label{fig:pointnet_loss_contours}
    \vspace{-1em}
\end{figure*}

Recent work on equivariant loss landscapes shows that equivariance constraints can interact with hidden parameter symmetries to create critical points and even spurious minima \cite{xie2025a}. Motivated by this view, we compare the local loss geometry around trained Adam and Muon models on ModelNet40. Following ~\citet{li2018visualizinglosslandscape}, we visualize loss slices around trained checkpoints. For each optimizer, we select the checkpoint whose accuracy is closest to the four-seed mean. Fig.~\ref{fig:pointnet_loss_contours} shows the PointNet loss slice, with corresponding 2D contours and 3D surfaces for the other architectures included in Appendix~\ref{app:loss_landscapes}.

The two optimizers lead to visibly different local geometry. In the PointNet example, the Adam slice has more irregular contour structure, while the Muon slice varies more smoothly across the plotted region. These plots are qualitative, since they only show a low-dimensional view of the full loss surface. This caution is especially relevant for equivariant models. \citet{xie2025a} show that hidden parameter symmetries can split an equivariant model's loss landscape into regions with different minima, so even if a different optimization method makes the landscape easier to traverse locally, it does not change the set of minima defined by the objective. Optimizer choice can nevertheless change the trajectory through the landscape, and therefore which solution is reached. The improvements in Table~\ref{tab:modelnet40} suggest that Muon can reach better solutions while keeping the equivariant architecture fixed. This is complementary to constraint-relaxation approaches, which modify the training problem to help SGD reach better minima \cite{pertigkiozoglou2024improving, manolache2025learning, elhag2025relaxed}. A plausible hypothesis is that Muon's orthogonalized momentum updates provide a trajectory-level bias toward more favorable regions of the landscape.

\begin{table}[t]
\caption{Representative Hessian estimates on clean ModelNet40 checkpoints.}
\label{tab:hessian}
\vskip 0.06in
\centering
{\setlength{\tabcolsep}{4.0pt}
\begin{tabular}{@{}llrrr@{}}
\toprule
Metric & Model & Adam & Muon & Ratio \\
\midrule
Top eig. & EGNN     & 27.14 & 128.83 & 4.75$\times$ \\
Top eig. & PointNet & 32.75 & 714.49 & 21.82$\times$ \\
Top eig. & DGCNN    & 12.14 & 136.23 & 11.22$\times$ \\
\midrule
Trace & EGNN     & 402.37 & 1472.78 & 3.66$\times$ \\
Trace & PointNet & 482.61 & 7362.05 & 15.25$\times$ \\
Trace & DGCNN    & 184.47 & 1218.74 & 6.61$\times$ \\
\bottomrule
\end{tabular}}
\end{table}

We next ask whether the smoother geometry corresponds to lower local curvature. For each trained checkpoint, we compute the loss on a fixed data subset and estimate Hessian summaries using autograd Hessian-vector products. We use power iteration for the top eigenvalue and a Hutchinson estimator with Rademacher probes for the trace \citep{pearlmutter1994fast,hutchinson1990stochastic,pmlr-v97-ghorbani19b}.

Table~\ref{tab:hessian} shows that the checkpoints reached by Muon have larger top-eigenvalue and trace estimates for every architecture. This rules out the most direct reading of the loss slices: the smoother contours are not evidence that Muon finds lower-curvature solutions. 
Around the Muon checkpoints, the sampled loss surface appears more regular, even though the curvature summaries at the checkpoint are larger.
This distinction is cautioned by \citet{dinh2017sharpminima}, who show that common notions of parameter-space sharpness can change under parameter symmetries and reparameterizations without changing the represented function. We therefore interpret the Hessian estimates and loss slices together: optimizer choice changes the local geometry of the reached checkpoint, but the change is not summarized by curvature alone.

\subsection{Spectral structure of weights and representations}

\begin{figure}
    \centering
    \includegraphics[width=.99\linewidth]{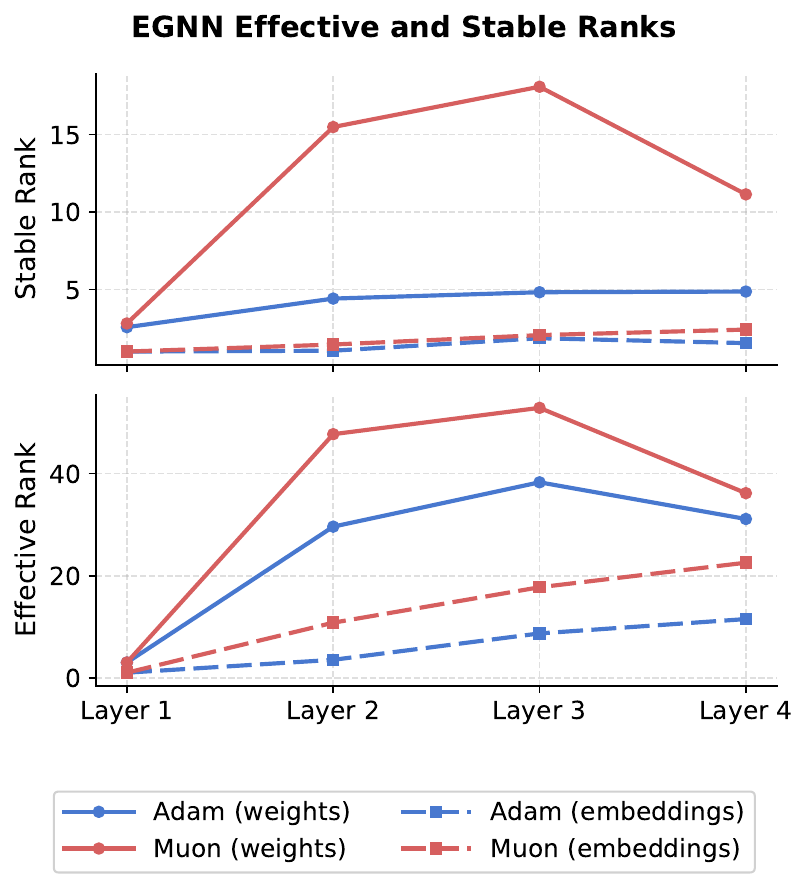}
    \caption{Per-layer stable rank (top) and effective rank (bottom) for EGNN trained on ModelNet40 with Adam (blue) and Muon (red). Solid lines are rank values for trainable weight matrices, with dashed lines from the corresponding intermediate embeddings. Muon produces less concentrated weight and activation spectra than Adam at every layer.}
    \label{fig:rank}
\end{figure}

For each trained checkpoint, we compute two spectral statistics on the singular values $\sigma_1 \geq \sigma_2 \geq \dots$ of every weight matrix $W$, and analogously of the representations at each layer. The first is the \textit{stable rank}, $\|W\|_F^2 / \|W\|_2^2 = \sum_i \sigma_i^2 / \sigma_1^2$; the second is the \textit{effective rank}, $\exp(H(p))$, where $p_i = \sigma_i / \sum_j \sigma_j$ and $H$ is the Shannon entropy~\citep{effrank}. Both lie in $[1, \mathrm{rank}(W)]$ and are maximized by a uniform spectrum, with smaller values indicating concentration along a few dominant singular directions. Throughout this section we analyze the trained Adam and Muon checkpoints from~\cref{sec:loss-geom}.

\paragraph{Weights.} Across all three ModelNet40 architectures, the checkpoints reached by Muon have higher stable and effective ranks than Adam at the majority of trainable layers. \cref{fig:rank} shows the EGNN case, where this holds at every layer. This contrasts with the low-rank implicit bias of gradient methods observed in other settings~\citep{arora2019}. Muon's orthogonalized momentum step instead treats singular directions of the update more evenly. \citet{jordan2024muon} motivate this design with the empirical observation that Adam and SGD-momentum updates for 2D parameters are often near-low-rank, with orthogonalization rescaling the ``rare directions'' that carry small but important update signal. The higher stable and effective ranks we observe in trained Muon weights are consistent with this picture.

\paragraph{Representations.} Low representation rank is a known failure mode in deep transformers, where pure self-attention layers with no skip connections drive outputs to rank one~\citep{Dong2021AttentionIN}, with associated gradient pathologies during training~\citep{noci2022}. The mechanism is attention-specific and our equivariant architectures differ in important ways, but it motivates examining representation rank in addition to weight rank. We extract activations at every layer of each trained model and compute the same spectral statistics on the resulting feature matrices, mean-pooling over points at intermediate layers and using each architecture's native pooling at the final layer. \cref{fig:rank} shows the per-layer comparison for EGNN, where the effect is most visible: Muon's representations stable and effective ranks are consistently higher than Adam's across the network, with roughly a $2\times$ effective-rank difference at the final layer. The same observations hold for DGCNN and PointNet (Appendix~\cref{app:ranks}), at different magnitudes. 

Both analyses thus point in the same direction at both the weight and the representation level, across architectures with quite different geometric inductive biases. We do not establish a causal link from these spectral properties to the accuracy gains in~\cref{tab:modelnet40,tab:qm9}, but the consistency of the effect motivates further investigation of how an optimizer's spectral behavior and a model's geometric constraints interact, and whether equivariant models in particular benefit from optimizers that distribute capacity across singular directions rather than concentrating it.

\section{Conclusions and further work}

We examined whether the optimizer, independent of architectural relaxations, can change the solutions reached when training equivariant networks. Muon improves over Adam on 3D point-cloud and most molecular tasks, and the trained Muon solutions differ from Adam's in their local loss geometry and in the spectral structure of their weights and intermediate representations. We see these results as preliminary: they identify the optimizer as a meaningful lever for equivariant model training, but the underlying mechanisms remain to be explained.

Several directions follow naturally. First, extending our post-hoc landscape and spectral analyses to training dynamics would show how the differences emerge during training, not only at convergence. Second, a more systematic study of symmetry groups, operations, and datasets is needed: in our experiments the largest gains appear on 3D tasks with $SE(3)$-style equivariance, while permutation-equivariant message passing on graphs benefits less. Third, the spectral results suggest an optimizer-design principle: update rules that keep the singular spectra of trained weights and representations from concentrating in a few directions may improve training in equivariant models. Muon provides one example through orthogonalized momentum updates, but other mechanisms could target this property more directly. More broadly, an optimizer designed for equivariance would exploit the structure of the symmetry group and constrained parameter space, rather than treating equivariance as a black-box constraint at training time.

\section*{Impact Statement}

This paper presents work whose goal is to advance the field of Machine
Learning. There are many potential societal consequences of our work, none
which we feel must be specifically highlighted here.

\bibliography{bibliography}
\bibliographystyle{icml2026}

\newpage
\appendix
\onecolumn
\section{More quantitative results}
\label{app:more_quantitative}

This section collects the additional graph message-passing experiments referenced in the main text. These experiments are intended to test whether the optimizer effect observed on the 3D ModelNet40 and QM9 settings transfers to permutation-equivariant models with graph benchmarks. The best single checkpoints improve under Muon, but the seed-averaged behavior is comparable to Adam.

For both datasets we use the same reporting convention as in the main text. Hyperparameters are selected separately for each optimizer by validation performance, four seeds are then trained with the selected configuration. We report both the best checkpoint observed across seeds and the mean performance across seeds. Peptides-func is evaluated by average precision, where larger is better, while ZINC is evaluated by mean absolute error, where smaller is better.

\begin{table*}[h]
\centering

\begin{minipage}{0.48\textwidth}
\centering
\caption{GINE performance on Peptides-func. Results are averaged over four seeds. Higher AP is better. Muon reaches the best checkpoint, while mean performance is comparable to Adam.}
\label{tab:pep-fun}
\begin{tabular}{lcc}
\toprule
Optimizer & Best AP $\uparrow$ & Mean AP $\uparrow$ \\
\midrule
Adam & 0.7098 & 0.7021{\tiny$\pm$0.0047} \\
Muon & \textbf{0.7201} & \textbf{0.7036}{\tiny$\pm$0.0108} \\
\bottomrule
\end{tabular}
\end{minipage}
\hfill
\begin{minipage}{0.48\textwidth}
\centering
\caption{GINE performance on ZINC. Results are averaged over four seeds. Lower MAE is better. Muon reaches the best checkpoint, while Adam has better mean performance.}
\label{tab:zinc}
\begin{tabular}{lcc}
\toprule
Optimizer & Best MAE $\downarrow$ & Mean MAE $\downarrow$ \\
\midrule
Adam & 0.1538 & \textbf{0.1692}{\tiny$\pm$0.0153} \\
Muon & \textbf{0.1425} & 0.1808{\tiny$\pm$0.0428} \\
\bottomrule
\end{tabular}
\end{minipage}

\end{table*}

The Peptides-func results from~\cref{tab:pep-fun} are close on average, but Muon obtains the strongest individual checkpoint, with the four-seed mean being only marginally higher than Adam, with larger variance.
On ZINC the picture is less favorable to Muon. Although Muon again reaches the best single checkpoint, its average MAE is worse than Adam's. These results support the observations made in the main text: changing the optimizer is clearly useful in the 3D geometric settings we study, but does not offer uniformly reliable improvement across all equivariance setups.

\section{Additional loss landscape visualizations}
\label{app:loss_landscapes}

\begin{figure*}[h]
    \centering
    \begin{subfigure}{0.48\textwidth}
        \centering
        \includegraphics[width=\linewidth]{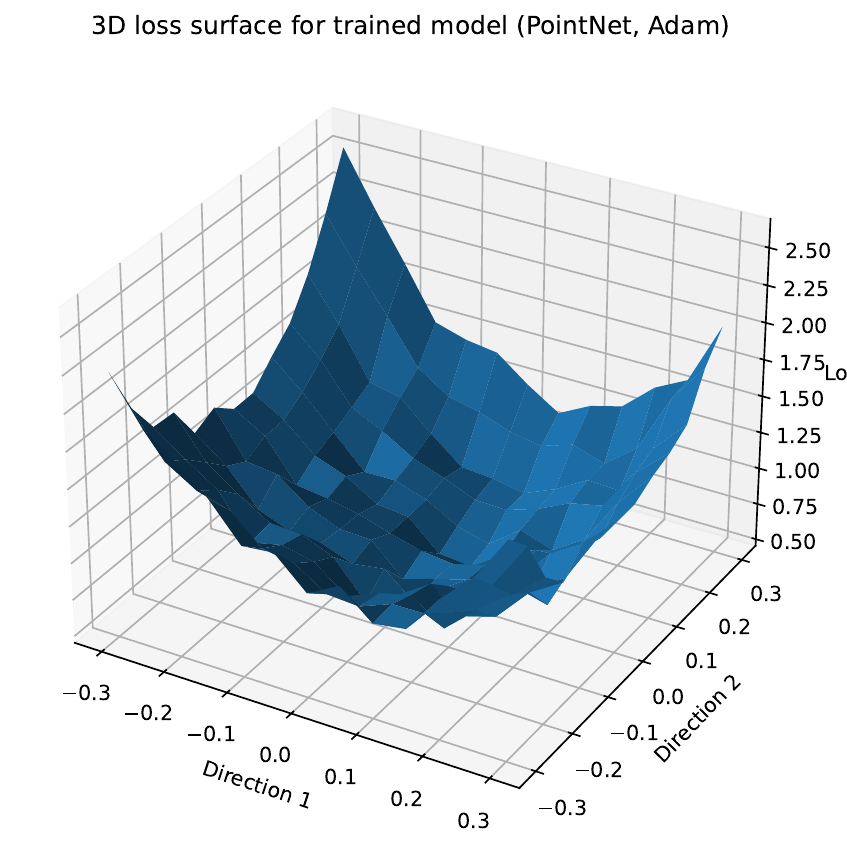}
        \caption{PointNet, Adam 3D surface}
    \end{subfigure}
    \hfill
    \begin{subfigure}{0.48\textwidth}
        \centering
        \includegraphics[width=\linewidth]{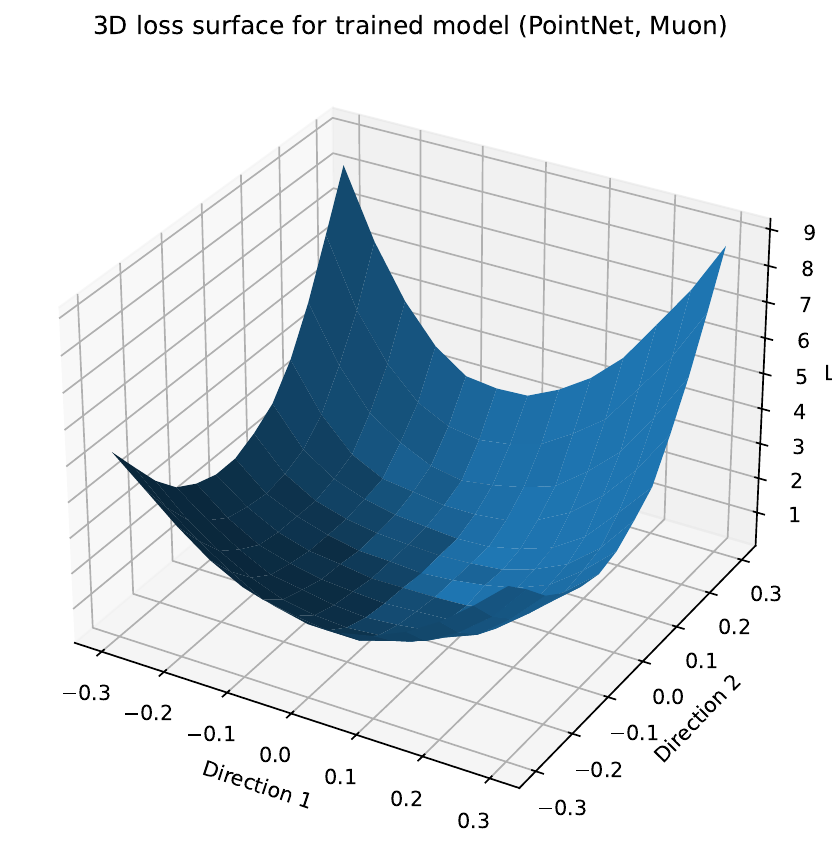}
        \caption{PointNet, Muon 3D surface}
    \end{subfigure}
    \caption{Additional PointNet 3D local loss surfaces around representative Adam and Muon checkpoints. These surfaces correspond to the same local slice construction used for the PointNet contours in the main text.}
    \label{fig:app_pointnet_loss_surface3d}
\end{figure*}

\begin{figure*}[h]
    \centering
    \begin{subfigure}{0.48\textwidth}
        \centering
        \includegraphics[width=\linewidth]{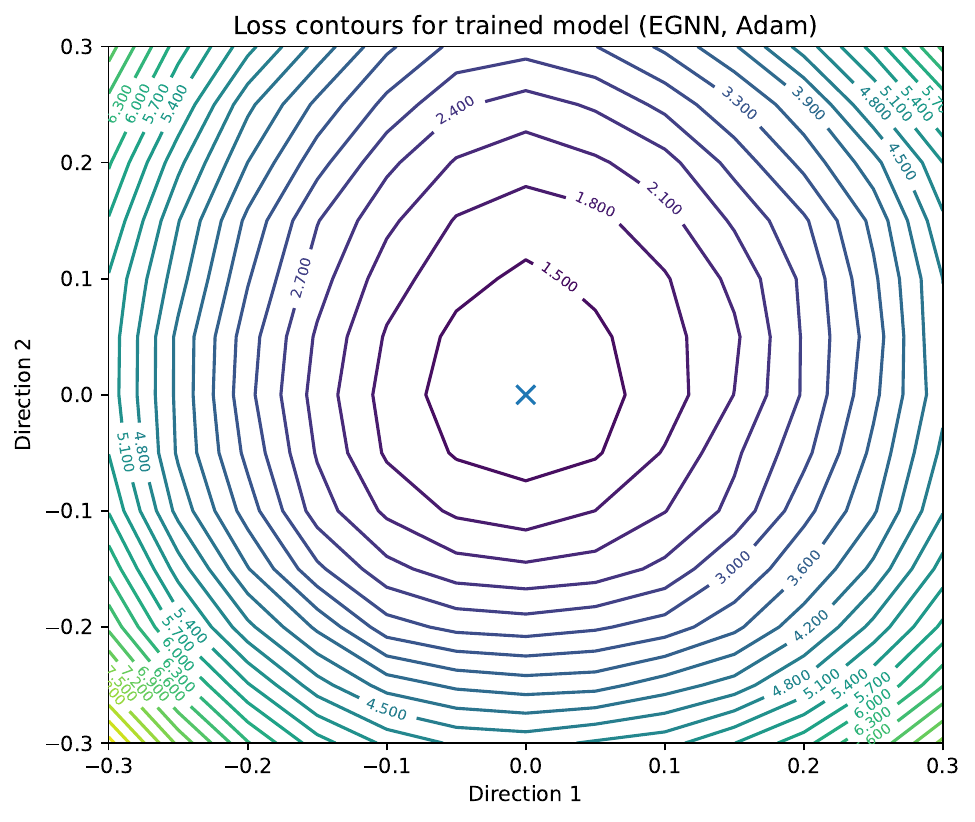}
        \caption{EGNN, Adam contour}
    \end{subfigure}
    \hfill
    \begin{subfigure}{0.48\textwidth}
        \centering
        \includegraphics[width=\linewidth]{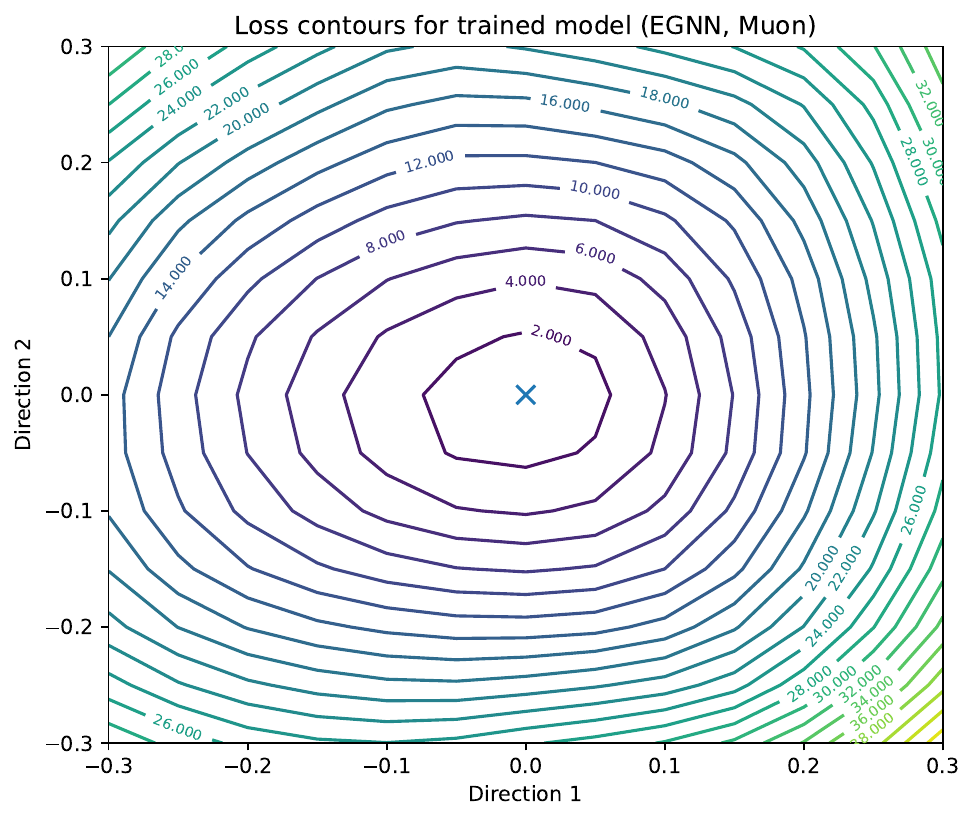}
        \caption{EGNN, Muon contour}
    \end{subfigure}
    \vspace{0.3em}
    \begin{subfigure}{0.48\textwidth}
        \centering
        \includegraphics[width=\linewidth]{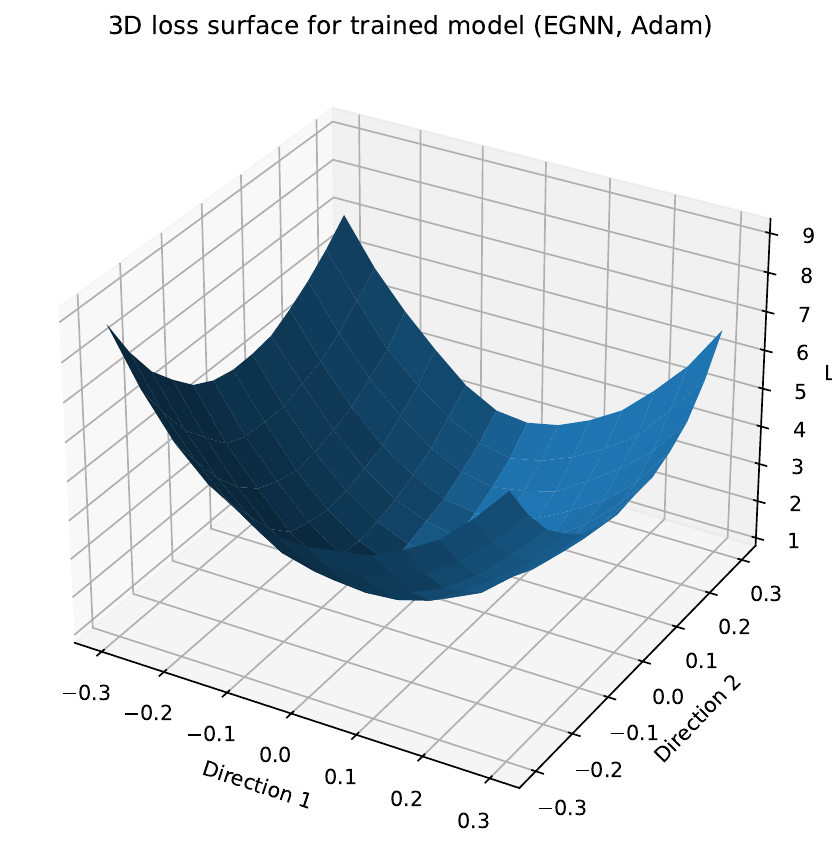}
        \caption{EGNN, Adam surface}
    \end{subfigure}
    \hfill
    \begin{subfigure}{0.48\textwidth}
        \centering
        \includegraphics[width=\linewidth]{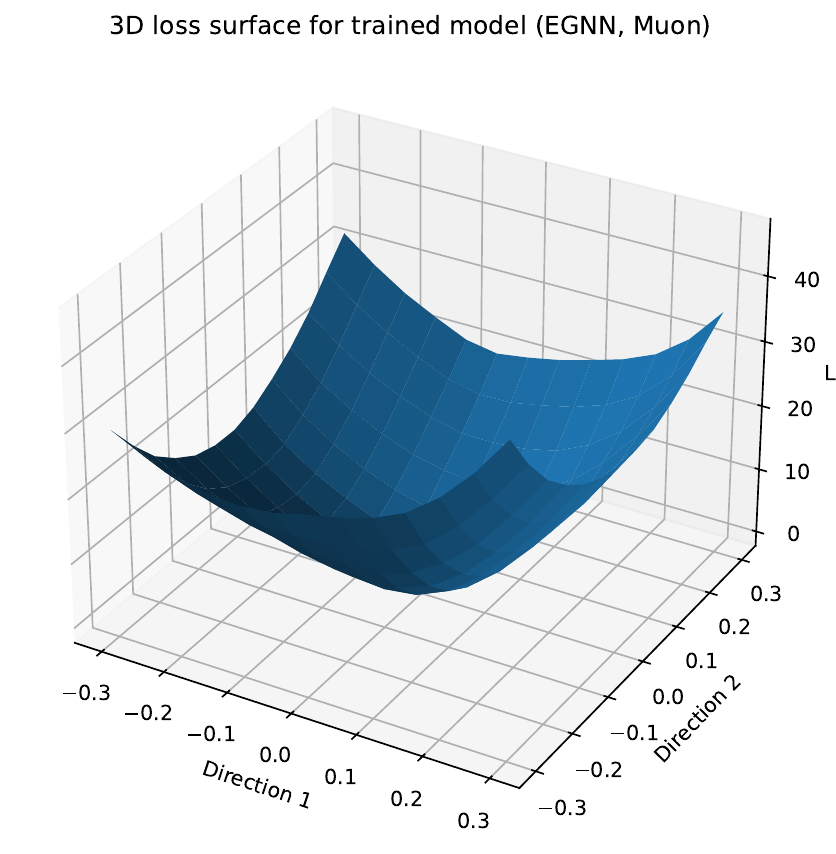}
        \caption{EGNN, Muon surface}
    \end{subfigure}
    \caption{Additional EGNN local loss visualizations around representative Adam and Muon checkpoints. The top row shows 2D contour slices and the bottom row shows the corresponding 3D loss surfaces.}
    \label{fig:app_egnn_loss_landscape}
\end{figure*}

\begin{figure*}[h]
    \centering
    \begin{subfigure}{0.48\textwidth}
        \centering
        \includegraphics[width=\linewidth]{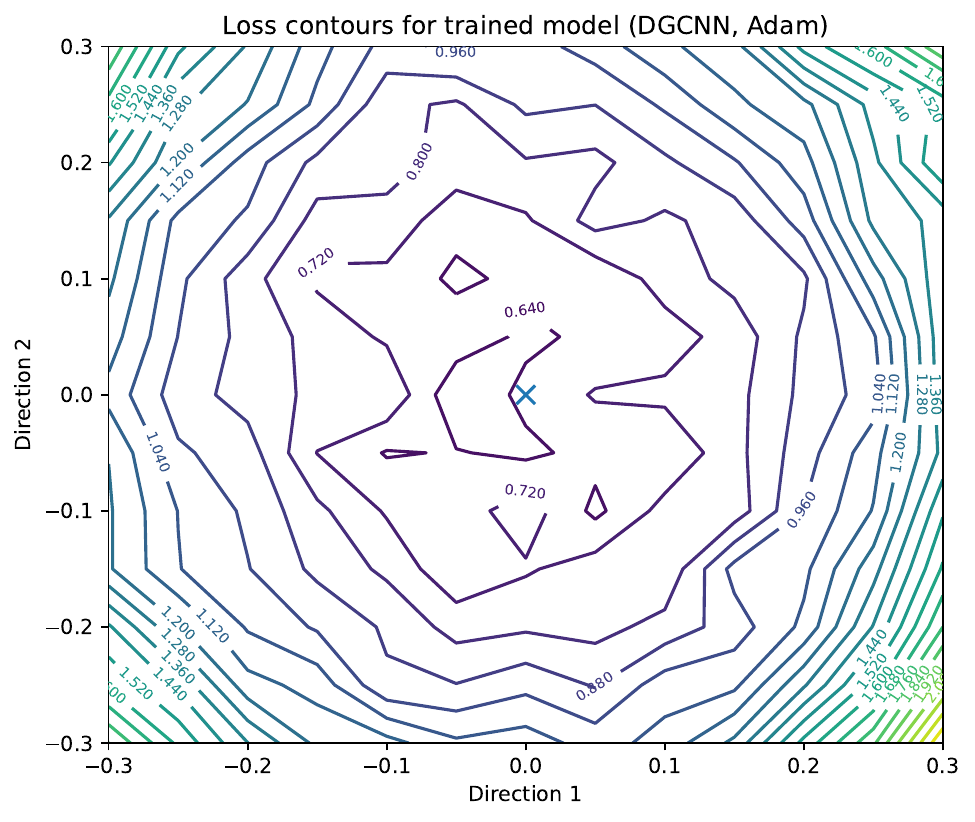}
        \caption{DGCNN, Adam contour}
    \end{subfigure}
    \hfill
    \begin{subfigure}{0.48\textwidth}
        \centering
        \includegraphics[width=\linewidth]{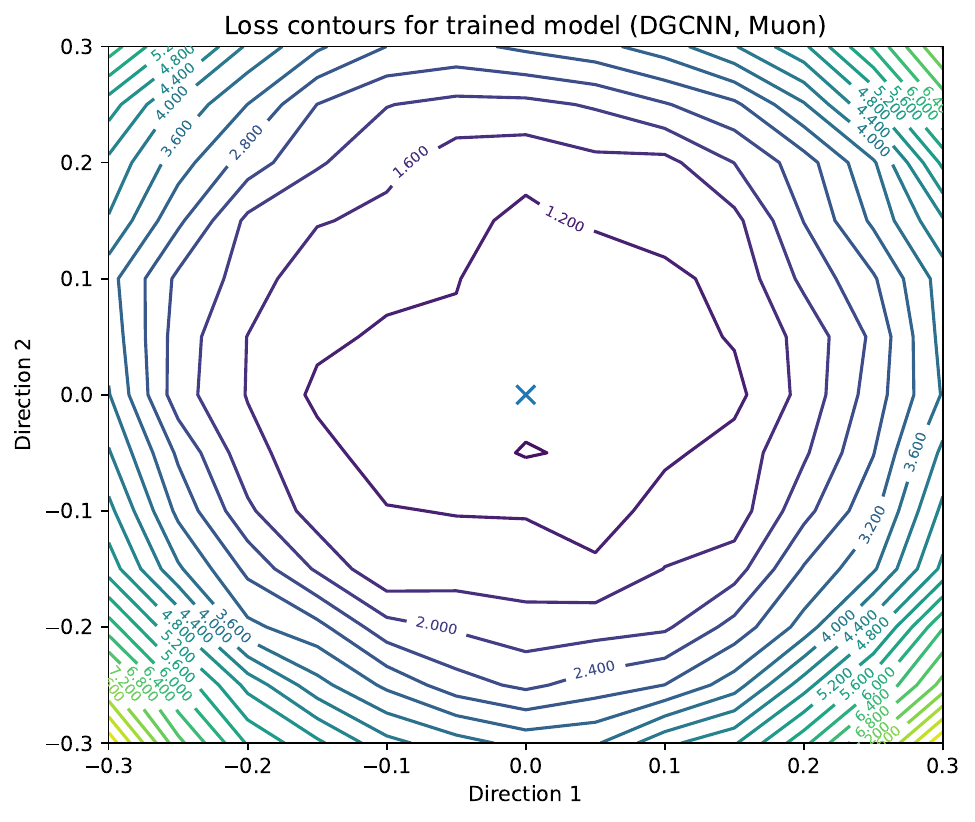}
        \caption{DGCNN, Muon contour}
    \end{subfigure}
    \vspace{0.3em}
    \begin{subfigure}{0.48\textwidth}
        \centering
        \includegraphics[width=\linewidth]{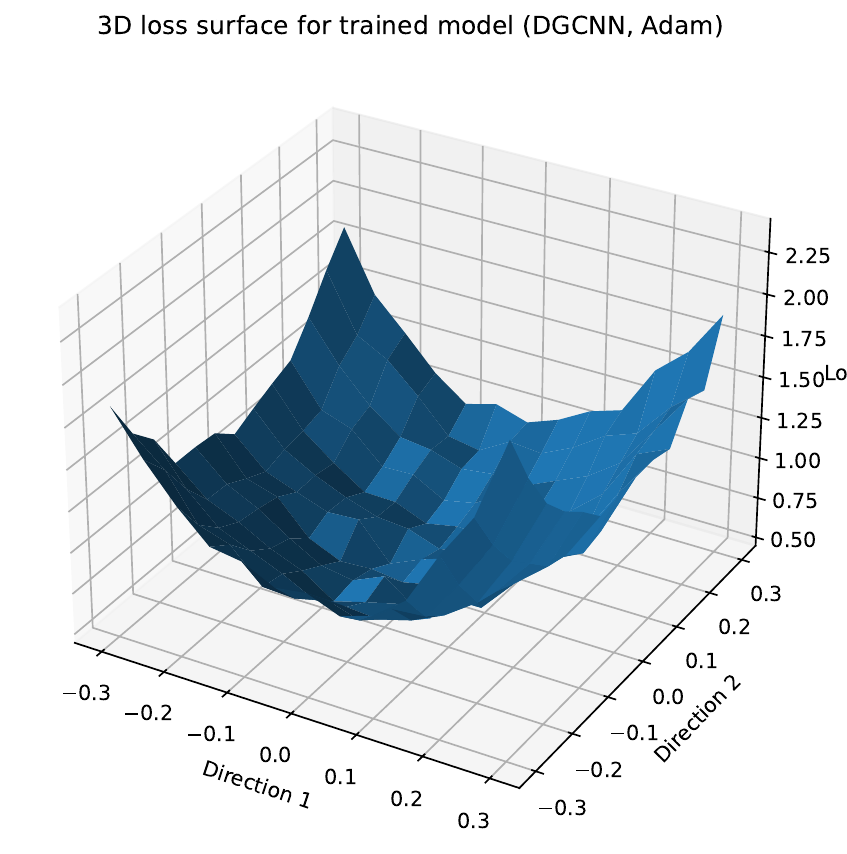}
        \caption{DGCNN, Adam surface}
    \end{subfigure}
    \hfill
    \begin{subfigure}{0.48\textwidth}
        \centering
        \includegraphics[width=\linewidth]{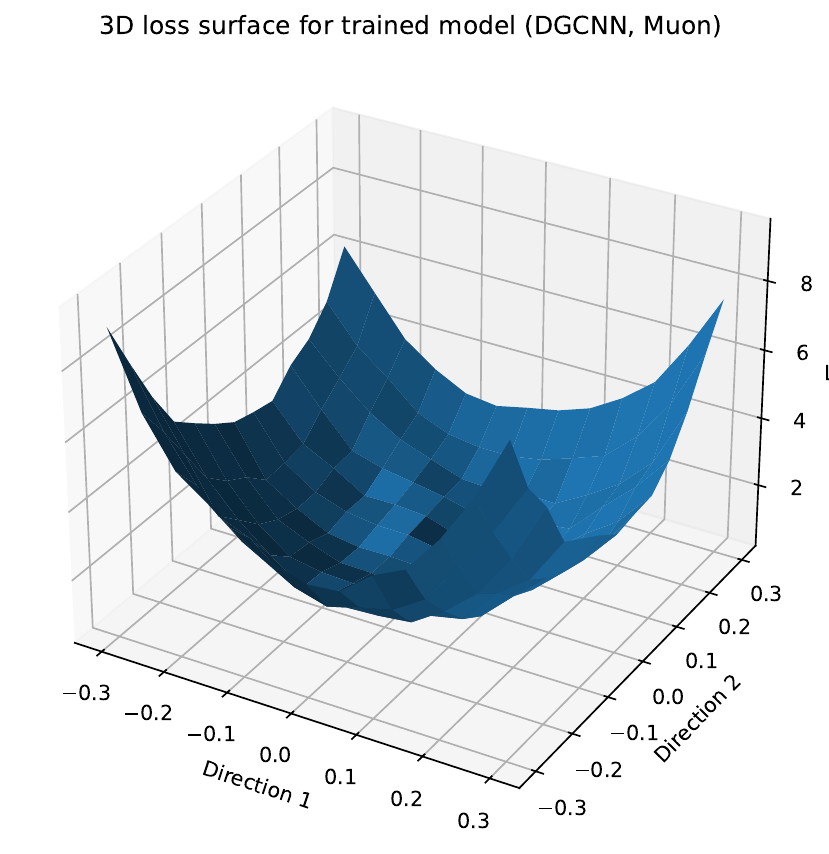}
        \caption{DGCNN, Muon surface}
    \end{subfigure}
    \caption{Additional DGCNN local loss visualizations around representative Adam and Muon checkpoints. The top row shows 2D contour slices and the bottom row shows the corresponding 3D loss surfaces.}
    \label{fig:app_dgcnn_loss_landscape}
\end{figure*}

\cref{fig:app_pointnet_loss_surface3d,fig:app_egnn_loss_landscape,fig:app_dgcnn_loss_landscape} extend the loss-landscape observations from the main text. They are included as qualitative local views around representative checkpoints. As in the PointNet comparison, each slice is centered at the selected trained checkpoint and uses normalized directions in parameter space. 

For DGCNN, the same pattern seen in the PointNet slice is visible at the level of these projections. The Adam slices contain more irregular contour geometry and sharper local variation, whereas the Muon slices are visually more regular over the plotted region. 

EGNN behaves differently. Both Adam and Muon produce visually regular slices over the plotted region, without the irregular contour structure visible in Adam's PointNet and DGCNN slices. The Hessian summaries in~\cref{tab:hessian} still place the Muon EGNN checkpoint at substantially higher top eigenvalue and trace than the Adam one, so the visual similarity does not extend to curvature at the checkpoint. We read this as additional support for the main-text observation that loss-slice plots and curvature summaries probe related, but complementary, and not identical properties of the local geometry.

\clearpage
\section{Additional effective ranks}
\label{app:ranks}

\begin{figure*}[h]
    \centering
    \begin{subfigure}{0.48\textwidth}
        \centering
        \includegraphics[width=\linewidth]{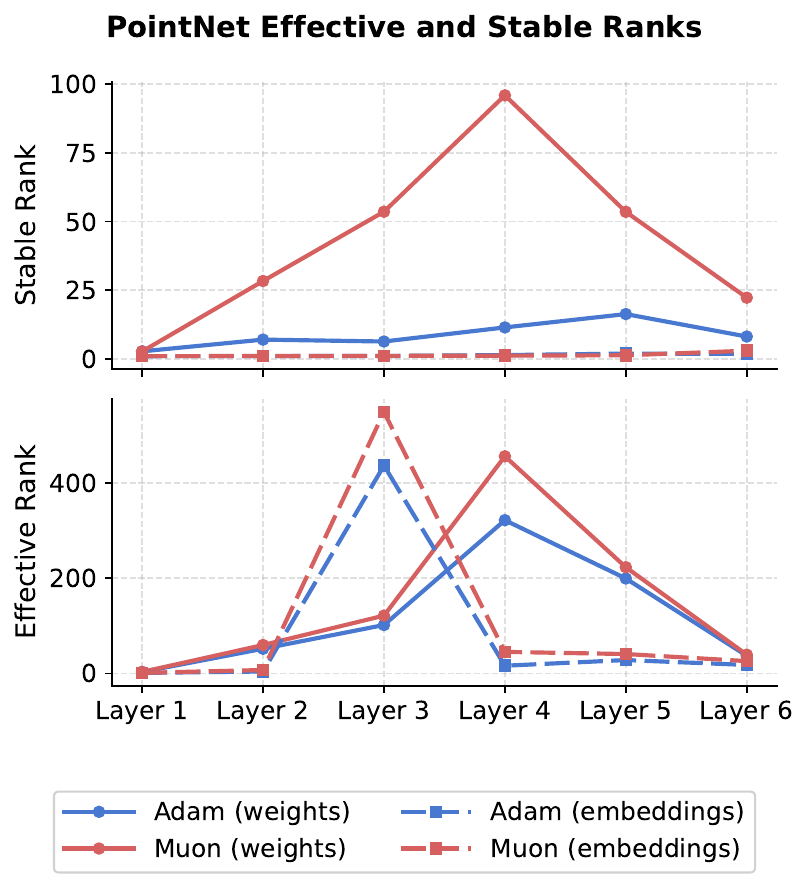}
        \caption{PointNet}
        \label{fig:app_pointnet_ranks}
    \end{subfigure}
    \hfill
    \begin{subfigure}{0.48\textwidth}
        \centering
        \includegraphics[width=\linewidth]{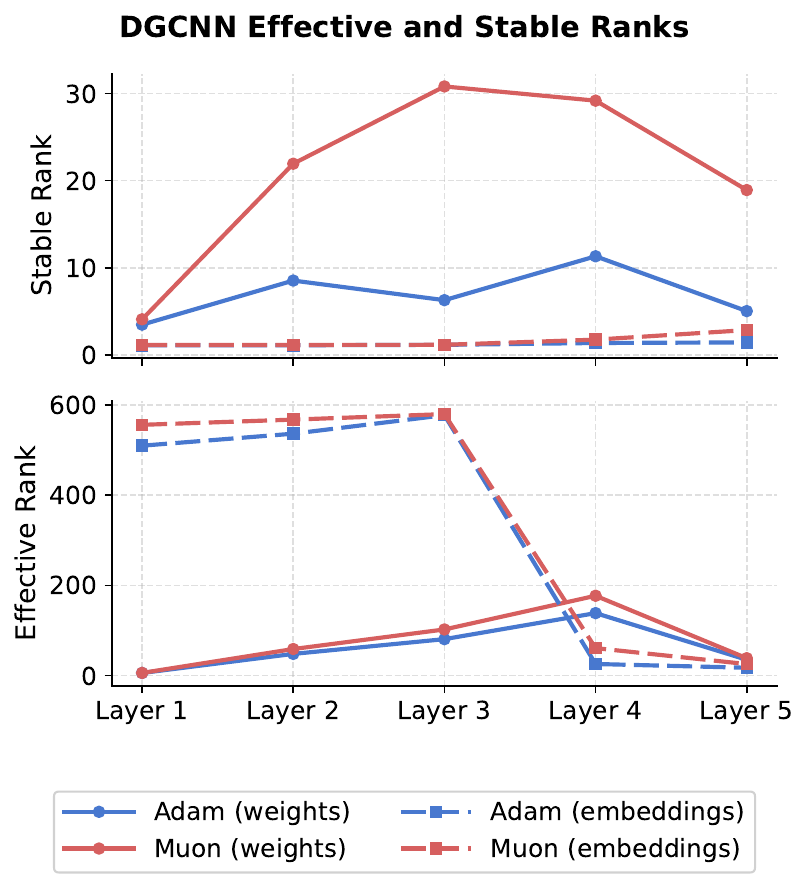}
        \caption{DGCNN}
        \label{fig:app_dgcnn_ranks}
    \end{subfigure}
    \caption{Additional per-layer stable and effective rank comparisons for weights and intermediate representations. Solid lines show trainable weight matrices and dashed lines show the corresponding representation matrices, computed on the fixed ModelNet40 test subset used in the main text.}
    \label{fig:app_additional_ranks}
\end{figure*}

\cref{fig:app_additional_ranks} shows the PointNet and DGCNN counterparts of the rank comparison reported for EGNN in the main text. The diagnostic is identical to the one used for EGNN: for trainable weight matrices and intermediate representations, we compute stable rank and entropy-based effective rank from the singular-value spectrum. Higher values indicate that the spectrum is less concentrated in a small number of dominant singular directions.

For PointNet, Muon increases the rank statistics for both weights and representations across most layers, with the largest visible separation in the weights stable rank, and in the effective rank curves. For DGCNN, the separation is smaller but follows the same direction.

\end{document}